\documentclass{article}

\usepackage[preprint]{neurips_2024}

\usepackage[utf8]{inputenc} % allow utf-8 input
\usepackage[T1]{fontenc}    % use 8-bit T1 fonts
\usepackage{hyperref}       % hyperlinks
\usepackage{url}            % simple URL typesetting
\usepackage{booktabs}       % professional-quality tables
\usepackage{amsfonts}       % blackboard math symbols
\usepackage{amsmath}        % for \eqref command
\usepackage{nicefrac}       % compact symbols for 1/2, etc.
\usepackage{microtype}      % microtypography
\usepackage{xcolor}         % colors

\usepackage{graphicx}
\usepackage{subcaption} 

\setcitestyle{semicolon}

\title{Auto-encoding Molecules:\\ Graph-Matching Capabilities Matter}
\author{%
    Magnus Cunow \\
  German Research Center for Artificial Intelligence (DFKI)\\
    Kaiserslautern \\
  \texttt{magnus.cunow@dfki.de} \\
   \And
   Gerrit Großmann \\
   German Research Center for Artificial Intelligence (DFKI) \\
   Kaiserslautern \\
   \texttt{gerrit.grossmann@dfki.de} \\
}

\begin{document}

\maketitle

\begin{abstract}
Autoencoders are effective deep learning models that can function as generative models and learn latent representations for downstream tasks. The use of graph autoencoders--—with both encoder and decoder implemented as message passing networks--—is intriguing due to their ability to generate permutation-invariant graph representations. However, this approach faces difficulties because decoding a graph structure from a single vector is challenging, and comparing input and output graphs requires an effective permutation-invariant similarity measure.
As a result, many studies rely on approximate methods. 

In this work, we explore the effect of graph matching precision on the training behavior and generation capabilities of a  Variational Autoencoder (VAE).
Our contribution is two-fold: (1) we propose a transformer-based message passing graph decoder as an alternative to a graph neural network decoder, that is more robust and expressive by leveraging global attention mechanisms. (2) We show that the precision of graph matching has significant impact on training behavior and is essential for effective \emph{de novo} (molecular) graph generation. % and representation learning.
%Tecnically, we dont do anything for representation learning
Code is available at {\href{https://github.com/mcunow/graph-matching}{\small \texttt{github.com/mcunow/graph-matching}}}.
\end{abstract}

 \section{Introduction}
One area where graphs are particularly well-suited is drug discovery, where they naturally represent molecular structures. This has led to a widespread use of graph-based deep learning methods for molecular prediction and generation. VAEs have emerged as powerful tools for molecular graph generation, enabling the exploration of the vast chemical space to identify drug-like molecules, while also facilitating molecular representation learning for downstream tasks~\citep{simonovsky2018graphvae,jin2018junction}. A graph autoencoder consists of an encoder that maps input graphs to a compact latent vector representation and a decoder that reconstructs the input graph from the latent vector~\citep{kipf2016variational}. 

This work addresses two of the main challenges of auto-encoding graphs.

First, decoding graphs from a single vector remains a challenging task due to the complex relationships between nodes and edges.
We address this limitation by proposing a graph transformer-based decoding approach, offering a more flexible and robust alternative to GNN-based decoders. While traditional methods typically rely on sequential models that suffer from limited flexibility and slow inference times, our approach leverages the power of transformers for efficient and expressive decoding.

Second, due to the lack of inherent node ordering, input and output cannot be directly compared entry-wise (as in image-based tasks).  
To assess reconstruction quality in a VAE's training objective, a \textit{permutation-invariant} loss is required. This can either use graph matching to establish explicit correspondences via a \textit{substitution matrix} for pointwise comparison or rely on indirect (potentially domain-specific) methods that trade granularity for efficiency.

VAEs often struggle with effective reconstruction, as small changes in the latent space, driven by their stochastic nature, can severely impact the output~\citep{muenkler2023vaes}. Therefore, we believe that a nuanced permutation-invariant loss, such as optimal graph matching, is essential for guiding the learning process.

In this work, we empirically investigate the impact of graph matching precision on the performance of auto-encoding molecular graphs.  
To this end, we compare input and output graphs using optimal and near-optimal graph matching. 
We demonstrate that graph matching quality significantly affects both the quality of the generated molecules and the convergence of the training loss.

\section{Method}

\subsection{Molecular Representation}
\begin{figure}[t]
    \centering
    \includegraphics[width=0.7\linewidth]{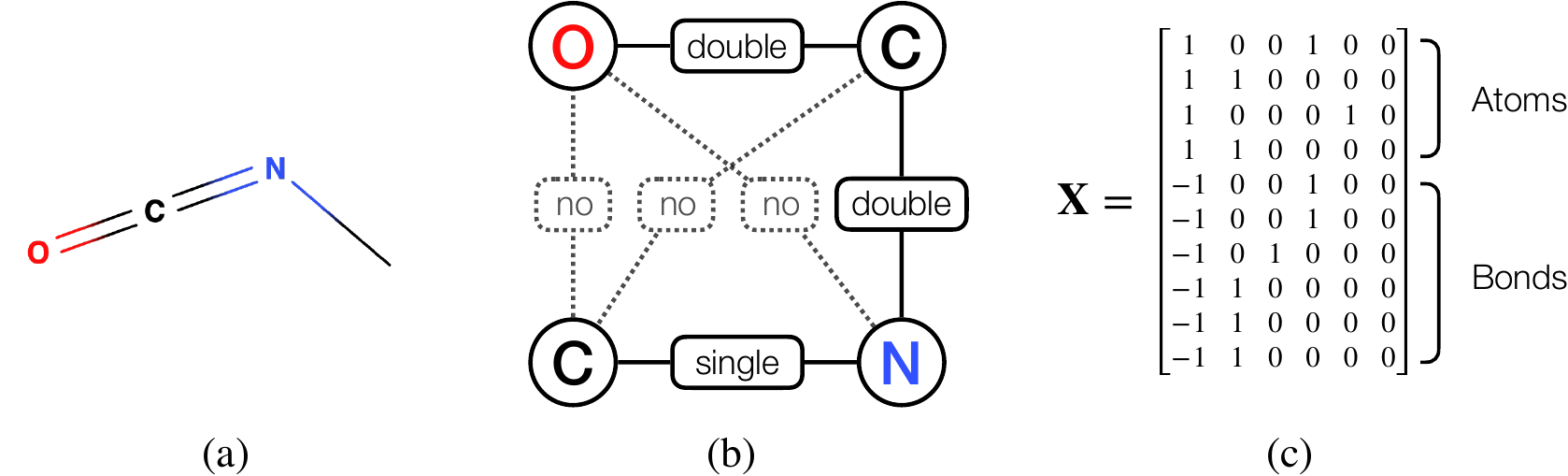}
    \caption{Schematic overview of our molecular representation. (a) Example molecule: \textit{methyl
isocyanate}. (b) Molecules are represented as fully-connected graph, where atoms, bonds and the absence of bonds between atoms are represented as nodes. Dotted lines indicate the introduction of edges for non-existing molecular bonds. (c) Our internal feature representation matrix \(X\), where the first column represents the type of the node (i.e., atom or bond). The remaining columns encode either the atom type (C, H, O, N, F) or the bond type (no bond, single, aromatic, double, triple). Source:~\cite{grossmanndiscriminator}.}
    \label{fig:mol_repr}
\end{figure}

A molecule can be represented as a graph, where atoms form nodes and bonds form edges, both of which can be featurized. In this work, we adopt a slightly different approach following~\citet{grossmanndiscriminator} that encodes atoms, bonds, and the \emph{absence} of bonds between atoms as graph nodes. This approach enables simultaneous node and edge convolutions, leveraging common message-passing techniques, which we can exploit in our decoder architecture.
A graph with \(n\) nodes is represented by a symmetric adjacency matrix \(A\in \{0,1\}^{n\times n}\) and a feature matrix \(X\in \mathbb{R}^{n \times d}\). The adjacency matrix encodes connectivity by linking \textit{atom}-nodes to \textit{edge}-nodes, where each \textit{edge}-node connects two \textit{atom}-nodes, as depicted in Figure~\ref{fig:mol_repr}. Additionally, we encode the absence of bonds between atoms, conceptually creating a quasi-fully-connected graph where each atom is connected to every other atom via an \textit{edge}-node. We omit hydrogen atoms and encode only atom and bond types as node features in \(X\).

\subsection{Variational Autoencoder}
A VAE is a generative model designed to learn a compact latent vector representation \(\textbf{z}\) of data \(\textbf{x}\) that captures the high-dimensional data distribution~\citep{kingma2013auto}. Autoencoders learn to compress data into a latent vector and then attempt to decode this vector back into the original input. We specifically examine their adaptation for learning graph representations~\citep{kipf2016variational}. Incorporating stochasticity into the latent space allows the model to capture the high-dimensional graph distribution and facilitates simple sampling from it. Framed as a variational inference problem, the optimization balances reconstruction quality with the enforcement of a latent space prior \(p(\textbf{z})\). This is achieved by computing a reconstruction loss \(L_{rec}\) to assess the similarity between the input graph \(G\) and the reconstructed, probabilistic graph \(\hat{G}\), and by applying a regularization penalty that aligns the latent space with a chosen prior. The regularization strategy thus ensures that the latent space is compact, continuous, and smooth.

\begin{figure}[t]
    \centering
    \includegraphics[width=0.6\linewidth]{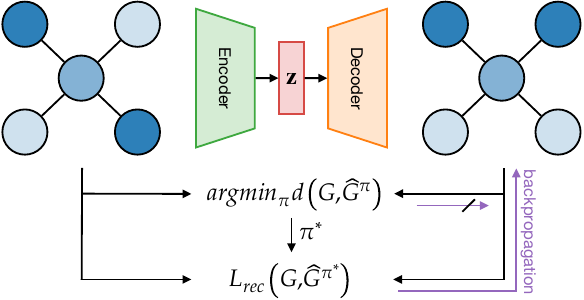}
    \caption{Permutation-invariant graph matching loss for graph autoencoders. The approach consists of two steps: (1) Compute the best alignment between the input graph \( G \) and the reconstructed probabilistic graph \( \hat{G} \) by finding the optimal permutation with respect to a given (not necessarily differentiable) distance measure \( d \). (2) Use the optimal permutation from the first step to compute the reconstruction loss \(L_{rec}\) and backpropagate gradients, as indicated by the purple arrow.}
    \label{fig:perm_loss}
\end{figure}

A graph autoencoder, as depicted in the upper part of Fig.~\ref{fig:perm_loss}, takes a graph \(G\) as input and consists of an encoder that learns a posterior \(q_\phi(\textbf{z}|G)\) to compress graphs into a dense vector and a decoder that defines the generative distribution \(p_\theta(G|\textbf{z})\) for reconstructing graphs from the latent space to a probabilistic graph \(\hat{G}\). The objective is to minimize the negative evidence lower bound (ELBO)

\begin{equation}
\mathcal{L}_{\mathrm{ELBO}}
= 
\underbrace{\;-\;\mathbb{E}_{q_\phi(\mathbf{z}\mid G)}\!\bigl[\log p_\theta(G\mid \mathbf{z})\bigr]}_{L_{\mathrm{rec}}}
\;+\;
\underbrace{D_{\mathrm{KL}}\!\bigl(q_\phi(\mathbf{z}\mid G) \,\|\, p(\mathbf{z})\bigr)}_{L_{\mathrm{reg}}}
\label{eq:elbo}
\end{equation}

on the marginal likelihood~\citep{kingma2013auto}. The first term corresponds to the reconstruction loss, which ensures that the decoded output resembles the input. The second term corresponds to the regularization term, enforcing that the learned latent distribution \(q_\phi(\mathbf{z}|G)\) remains close to the prior \(p(\mathbf{z})=\mathcal{N}(0,I)\), which is chosen to be a isotropic Gaussian.

For graph-based VAEs it is widely accepted that encoding can be efficiently handled by GNNs. However, decoding remains a more complex challenge, with no universally agreed-upon approach to date.

We build a VAE that is fully permutation-invariant, ensuring that the latent representation comprises only permutation-invariant features and that the reconstruction loss \(L_{rec}\) is independent of the internal node ordering. In other words, any permutation of the reconstructed input yields the same \(L_{rec}\). To achieve this, we employ a permutation-invariant encoder so that only permutation-invariant features are learned and encoded into the latent space. Additionally, we design a decoder that randomly determines the node ordering, and we evaluate the reconstruction quality using a permutation-invariant loss. Together, these components guarantee that the model learns representations and reconstructions that are invariant to node permutations.

\subsubsection{GNN Encoder}
The encoder is built on a Graph Isomorphism Network (GIN), to balance expressiveness and efficiency~\citep{xu2018powerful}. The GIN computes permutation-equivariant node embeddings that are subsequently aggregated into a single graph-level representation using an attention-based aggregation mechanism~\citep{li2019graph}. This process yields a permutation-invariant global latent representation, ensuring that the encoder's output remains independent of node ordering.

\subsubsection{Transformer Decoder}
We employ a one-shot, graph transformer–based architecture for the decoder---a novel approach to our knowledge. Specifically, we utilize a transformer model that leverages attention mechanisms for interactions between tokens~\citep{vaswani2017attention}. The decoder simultaneously constructs both node features and edge types while ensuring that the initialization respects the number of input nodes---a permutation-invariant property of a graph.

We initialize tokens representing a fully-connected graph (i.e., each node and each edge is represented by a token), with each token seeded by the global latent graph representation. Following the approach of~\citet{kim2022pure}, we assign each token a type identifier (node or edge) and use orthogonal random features as relative positional encodings that capture connectivity between nodes. This initialization strategy enables the decoder to infer the underlying graph structure while ensuring that no absolute positional information is available to any token. As a result, the decoder is both permutation-equivariant and non-deterministic. It is important to note that this does not imply an absence of ordering during the expansion process; rather, it means that the transformer cannot explicitly determine the absolute position of any token.

When combined with a permutation-invariant loss, this approach allows the decoder to autonomously determine the node ordering. The key idea is that node ordering is neither encoded into the latent space nor dictated by the input dataset. Instead, the latent vector provides a global context to guide reconstruction, while the transformer's attention mechanism learns the relationships between tokens, modeling the interdependencies among atoms and bonds. Consequently, the node ordering is determined randomly during the decoding process.

Our early experiments revealed that GNN-based decoders struggle with decoding nodes and edges simultaneously—a limitation also noted in the sparse literature on GNN-based decoders. In contrast, our transformer-based decoder effectively handles the concurrent processing of node and edge features, leveraging their correlations. This capability is in line with the few existing one-shot decoder approaches that address similar challenges~\citep{flam2020graph}.

\subsection{Similarity Measure}
Graphs represent relational data which inherently models relationships between a set of unordered entities. This has several severe consequences when processing graph data, as it requires setting an internal node order, which is the reason why GNNs' permutation equivariance imposes an inductive bias on internal graph representations. Unlike images, graphs cannot be directly compared pointwise due to their arbitrary node ordering, which can lead to significantly different internal representations.
Hence, comparing two graphs \(G_1\) and \(G_2\) with \(|V_1|=|V_2|\) and arbitrary node ordering, where node correspondence is unknown, requires a similarity criterion \(S(G_1,G_2)\) that is permutation-invariant 
\[ S(G_1,G_2)=S(G_1^{\pi_1},G_2^{\pi_2}) \quad \forall \pi_1,\pi_2 \in P,\]
where \(P\) is the set of all \(n!\) permutations for a graph with \(n\) nodes. In this work, we consider similarity only on a domain-agnostic level with respect to the graph's connectivity, features, and structure. Generally, there exists a trade-off between granularity and complexity for similarity measures. Simple, yet effective, permutation-invariant similarity criteria include graph statistics such as node, edge, or degree distributions; however, their capability to distinguish between similar graphs is very limited. More advanced techniques, such as the graph edit distance—which measures the minimum number of operations required to transform one graph into another—provide a more nuanced measure but are not differentiable and hence are not suitable as a loss function for deep learning methods.

Molecules are very sensitive structures that adhere to chemical constraints. A more fine-grained approach provides nuanced feedback, helping the model learn these underlying chemical rules and decode latent graph representations. Hence, a strong similarity measure enables a model to exploit the expressivity of the measure and leverage its own capacity to learn more meaningful molecular representations by improving the reconstruction of the input. We use graph matching to define a permutation-invariant similarity measure that is differentiable and consequently serves as a reconstruction loss for autoencoders, which is depicted in Figure~\ref{fig:perm_loss}.

\subsubsection{Graph Matching and Reconstruction Loss}
We define our differentiable reconstruction loss function using
graph similarity measure that is based on graph matching.
The objective of graph matching is to find a correspondence 
or substitution matrix between the nodes of the features of the input and output graphs.
We consider only bijective (i.e., \emph{one-to-one}) correspondences and define it solely for \(|V_1|=|V_2|=n\). 

Specifically, we define the reconstruction loss from Equation \eqref{eq:elbo} using a \emph{permutation} or \emph{substitution matrix} $\pi$ as  

\[
    L_{\mathrm{rec}}(G_1, G_2) = \|X_1 - X_2^{\pi}\|^2,
\]

where $X_1$ is the matrix representation of the input graph, $X_2$ is the matrix representation of the output graph, and $X_2^{\pi}$ is the permuted matrix representation of the output graph. Since we use the Frobenius norm, this loss is proportional to the mean squared error.

The reordering of \emph{bond} nodes is entirely determined by the reordering of \emph{atom} nodes, ensuring that all edge relations remain consistent. In our molecular graph, any given permutation applies directly to the \emph{atom} nodes, while the \emph{bond} nodes are subsequently adjusted to preserve their connectivity. This guarantees that, conceptually, only the \emph{atom} nodes undergo reordering, while the \emph{bond} structure is maintained within the transformed graph.

\paragraph{Optimal Graph Matching.}  
When computing $L_{\mathrm{rec}}$, we need to choose a permutation $\pi$.
We refer to an \emph{optimal graph matching} if $\pi$ is chosen as $\pi^*$, the permutation that minimizes the reconstruction loss.  
That is,  

\[
\pi^* = \arg\min_\pi d(X_1, X_2^\pi) = \arg\min_\pi \|X_1 - X_2^\pi \|^2 \,.
\]

Note that in this work, we use the mean squared error to determine the best permutation and compute the reconstruction loss.  
Technically, $\pi$ could be chosen to minimize a distance function $d$ that serves only as a surrogate for the actual reconstruction loss.

\paragraph{Near Optimal Graph Matching.}
In practice, it is typically, not feasible to find the optimal $\pi$.
A graph matching approach that does not find the global minimum (i.e., $\pi \neq \pi^*$) is considered as less precise.
In this work, we test how choosing $\pi$ among the top-$k$ permutations influences training.

%Finding the best alignment between graphs is expensive, hence one might be tempted to use less expressive distance measures or approximate the optimal correspondence. 
Specifically, we argue that finding the optimal correspondence critically impacts gradient signals, training stability, convergence and performance. The argument is that for isomorphic graphs (and w.l.o.g.\ other graph pairs), an optimal matching yields zero loss, while suboptimal or poor matching introduces undesirable gradient noise and higher loss. Thus, good graph matching is essential for robust training dynamics.

Finding a good correspondence becomes significantly more difficult for larger graphs and remains an open research field, therefore we focus on the effects of graph matching for training an autoencoder on small graphs.

\section{Results}

We trained multiple models on the QM9 dataset by~\citet{ramakrishnan2014quantum} using the described permutation-invariant loss.
\paragraph{Setup.}
The model was evaluated by periodically generating 10,000 molecules and determining the validity (fraction of graphs that resemble valid molecules), uniqueness (fraction of valid molecules that were generated once), and novelty (fraction of unique molecules that are not in the dataset) using RDKit. These metrics are commonly used to indicate how well the model has learned the molecular distribution. However, they do not account for synthesizability or drug-likeness, and models might optimize for these metrics, thereby limiting their explanatory power. 
The experimental setup included augmenting the graph view during training by random permutations to ensure that the model does not learn any rules of input ordering.
\paragraph{Implementation Details.}
We implemented the model using PyTorch and PyTorch Geometric. The encoder is a five-layer GIN architecture with hidden dimensions \([16,32,64,128,256]\), resulting in a total of 133,861 trainable parameters. The latent space has a dimensionality of 50.
The decoder follows a transformer encoder architecture with five layers, each having a hidden dimension of 128 and eight attention heads. In total, the decoder has 3,093,864 trainable parameters. Both the encoder and decoder incorporate layer normalization. Further implementation details and hyperparameters can be found in the repository.

\paragraph{Top-\(k\) Matching.}
We compared the optimal graph matching approach with several near-optimal alternatives by sampling from a predefined range of top permutations. Specifically, we trained multiple models, each sampling from a set of permutations: \textit{top 10}, \textit{top 50}, and \textit{top 100}, to simulate near-optimal graph matching. The optimal matching corresponds to the \textit{top 1} permutation, representing the best possible match.
%For robustness, we sampled multiple graph embeddings per update. 

\paragraph{Baselines.}

To compare the optimal and near-optimal matching, we used three baseline approaches:

\begin{itemize}
    \item \textbf{No Matching:} The simplest approach where no matching is applied.
    \item \textbf{Statistics:} A method that relies on (permutation-invariant) global graph statistics (such as feature and degree distribution) without enforcing node-level matching.
    \item \textbf{GNN:} A GNN-based approach that computes and compares (permutation-invariant) graph embeddings of the two graphs.
\end{itemize}

\begin{figure}[t]
    \centering
    \begin{subfigure}[t]{0.48\textwidth} % Force same width
        \centering
        \includegraphics[width=\linewidth]{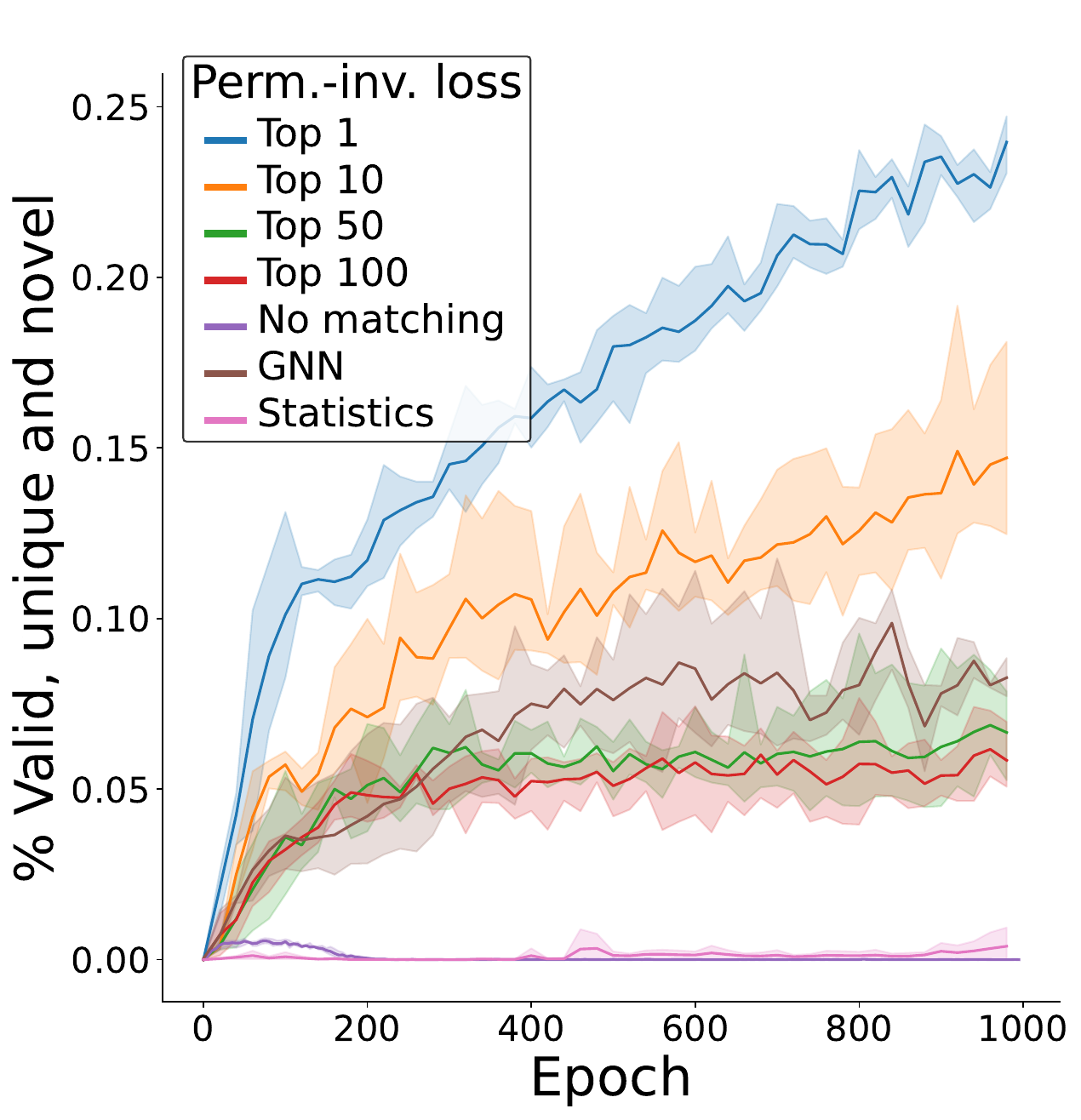}
        \caption{Fraction of valid, unique, and novel generated molecular graphs.}
        \label{fig:metric}
    \end{subfigure}
    \hfill
    \begin{subfigure}[t]{0.48\textwidth} % Force same width
        \centering
        \includegraphics[width=\linewidth]{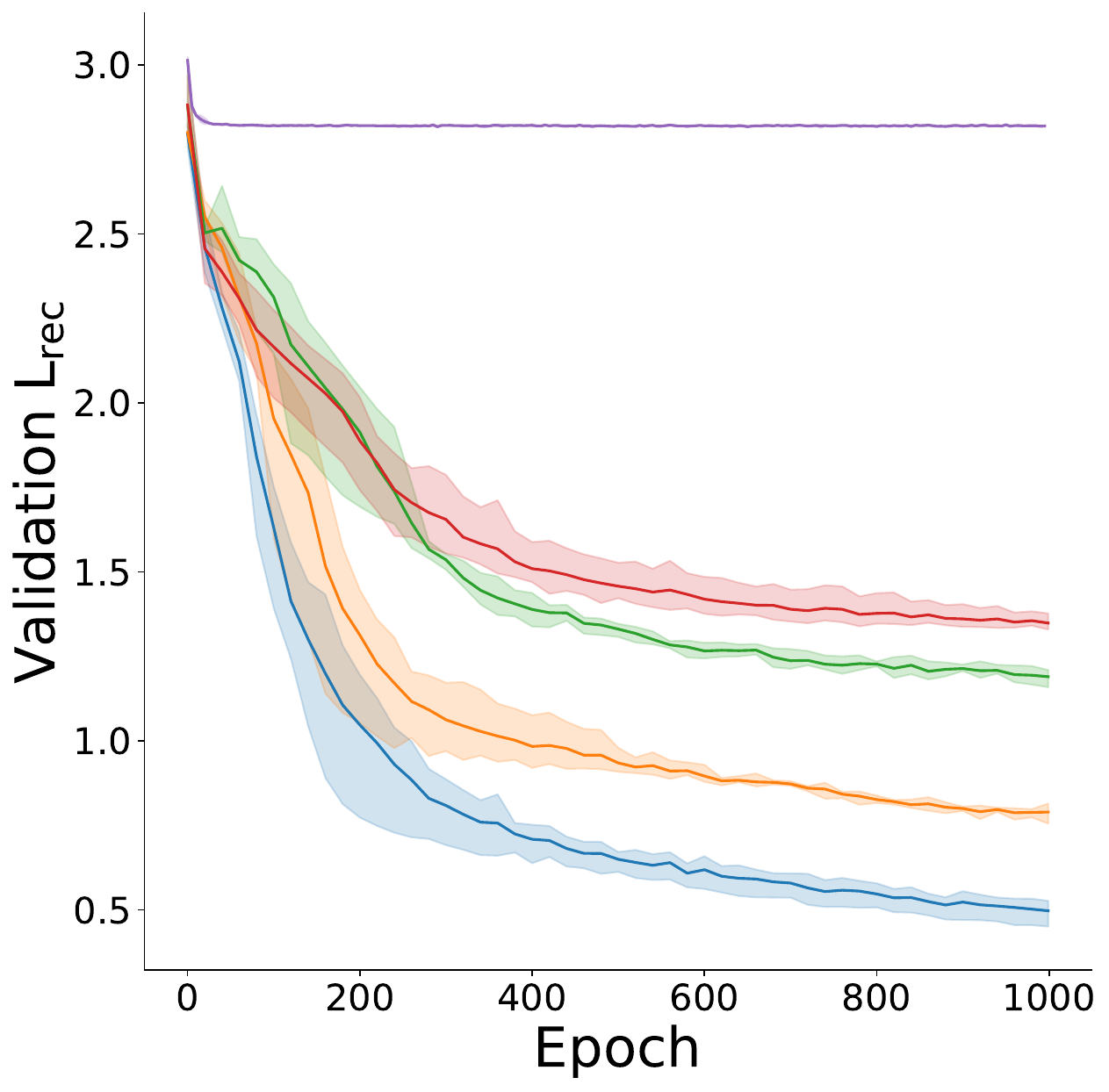}
        \caption{Validation reconstruction loss.}
        \label{fig:val}
    \end{subfigure}
    \caption{Comparison of training behavior and generation capabilities, evaluated by graph matching quality. A baseline is also included, using no matching at all, a simple graph statistics loss and a GNN-based loss. %(a) Fraction of valid, unique, and novel generated molecular graphs. (b) Validation reconstruction loss.
    }
    \label{fig:combined}
\end{figure}

\paragraph{Training Behavior.}
In Figure~\ref{fig:val}, we examine the reconstruction loss on the validation set across five models. For comparability, we focus only on the graph matching approaches. Our results show that the absence of graph matching leads to immediate convergence to a sub-optimal solution. Conversely, optimal graph matching produces the steepest learning curve, and we observe that the higher the precision of graph matching, the steeper the learning curve becomes.

\paragraph{De Novo Molecule Generation.}
In Figure~\ref{fig:metric}, we evaluate our model based on the overall fraction of valid, unique, and novel molecules generated. These metrics are interdependent, as poor performance on any one of them creates a bottleneck. Unlike many studies, we did not apply search algorithms to refine and sanitize the generated molecules before evaluation. Our results demonstrate that optimal graph matching consistently improves these metrics, while near-optimal matching shows inconsistent improvements but still captures important features of the distribution. These findings provide strong evidence that better graph matching enhances reconstruction, enabling the model to learn more nuanced features of the underlying distribution.

In Table~\ref{table:evaluation}, we summarize the best results from the five models based on our evaluation metrics. The results indicate that optimal graph matching achieves the overall best performance. Notably, graph matching significantly improves the uniqueness score, albeit with a slight decrease in validity. This suggests that the model has learned more nuanced representations but has not fully captured the chemical constraints required for valid molecule generation. As indicated by Figure~\ref{fig:val}, this limitation may be addressed with longer training periods, as the model appears not to have fully converged.

\begin{table}
  \caption{Evaluation of validity, uniqueness, and novelty of generated molecules between best models.}
  \label{table:evaluation}
  \centering
  \begin{tabular}{lllll}
    %\toprule
    %\multicolumn{2}{c}{Part}\\
    \cmidrule(r){1-5}
    Matching     & Validity     & Uniqueness  & Novelty  &  Overall fraction \\
    \midrule
    No matching        & 1      & 0     & 0     & 0\\
    GNN                & 0.41   & 0.21  & 0.93  & 0.08\\
    Top 100            & 0.83   & 0.13  & 0.62  & 0.07\\
    Top 50             & 0.80   & 0.16  & 0.68  & 0.09\\
    Top 10             & 0.80   & 0.33  & 0.68  & 0.18\\
    Top 1              & 0.70   & 0.48  & 0.74  & 0.25\\
    \bottomrule
  \end{tabular}
\end{table}

\section{Related Work}
Graph-based molecular representation learning has advanced significantly, with methods typically categorized into two areas: property prediction and molecular generation. Property prediction focuses on estimating molecular or atomic properties, while generation methods—including VAEs, normalizing flows, diffusion models, and GANs—compare input and output during training in some form, except for GANs. Among these approaches, VAEs have been widely explored for molecular generation, offering a probabilistic framework that balances reconstruction quality and latent space regularization.

\paragraph{Graph-Based Generation.}  
A wide range of graph-based VAEs have been proposed, utilizing either one-shot or sequential decoding to reconstruct molecular graphs. Sequential decoding has been extensively studied and incorporates strategies such as molecular scaffolds~\citep{jin2018junction,li2018multi,lim2020scaffold}, motif-based learning~\citep{jin2020hierarchical,maziarz2021learning,gao2023fragment}, and reinforcement learning~\citep{kearnes2019decoding}. Masking techniques have also been employed to guide decoding more effectively~\citep{samanta2020nevae}. In contrast, one-shot decoding, though conceptually appealing, has received less attention due to challenges in stability and convergence. Efforts to improve one-shot decoding have included MLP-based feature prediction~\citep{simonovsky2018graphvae,bresson2019two}, message-passing decoders~\citep{flam2020graph}, and hybrid architectures that combine both approaches~\citep{bresson2019two}.\\
Beyond VAEs, normalizing flow-based methods have demonstrated promise in graph generation, leveraging invertible transformations to learn complex data distributions~\citep{madhawa2019graphnvp,shi2020graphaf,zang2020moflow}. More recently, diffusion models have gained attention for molecular generation, with equivariant formulations showing advantages in data efficiency and parameter efficiency~\citep{hoogeboom2022equivariant,jo2022score,vignac2022digress,brehmer2024does}. Unlike VAEs, diffusion and normalizing flow models inherently handle permutation invariance, circumventing the need for explicit graph matching.

\paragraph{Reconstruction.}  
In contrast, VAEs rely on a reconstruction loss to evaluate generated graphs, introducing the challenge of permutation invariance. Early work, such as~\citet{kwon2019efficient}, used simple graph statistics for evaluation, while scaffold-based methods introduced constraints to simplify graph matching~\citep{lim2020scaffold,jin2018junction}. Other approaches have employed search algorithms to enforce node ordering during sequential decoding~\citep{jin2020hierarchical} or incorporated reinforcement learning to embed permutation-invariant information in the reward function~\citep{kearnes2019decoding}. Some methods attempt to circumvent the need for explicit permutation invariance by designing decoders that naturally align nodes without requiring a specialized loss function~\citep{flam2020graph}.

\paragraph{Graph Matching.}
Graph matching remains a central issue in evaluating reconstruction quality. Various techniques have been proposed, ranging from max-pooling-based alignment~\citep{simonovsky2018graphvae} to the use of GNNs for computing permutation-invariant losses. Optimal transport-based methods~\citep{tang2022graph,gasteiger2021scalable,shervashidze2011weisfeiler,pellizzoni2024structure} offer principled solutions, while alternative techniques attempt to approximate non-differentiable distance measures~\citep{bai2019simgnn}. Despite these efforts, the problem of evaluating reconstruction loss in permutation-invariant models remains largely unresolved.

%\paragraph{Open Challenges.}  
%While significant progress has been made, a critical gap persists in %the formulation and evaluation of permutation-invariant reconstruction losses. 

\section{Conclusion}
In this work, we investigated the impact of graph matching on the ability of VAEs to autoencode molecular graphs. Our findings demonstrate that the precision of graph matching is a crucial factor for both generation quality and training convergence.

In a broader context, we believe this insight extends to larger graphs and other complex domains where point-wise correspondence cannot be directly computed.

To fully leverage graph matching, we introduced a powerful transformer-based decoder as an alternative to one-shot GNN-based decoders. This decoder enables simultaneous convolution over both node and edge features while incorporating an effective global attention mechanism to guide the construction process. However, these advantages come with challenges, including higher data requirements and less favorable scaling behavior.
We believe future work can address these limitations by improving data efficiency and scalability.

Most importantly, our method is currently limited to small graphs due to the combinatorial explosion of possible permutations as the number of nodes increases. Consequently, for larger graphs, the use of heuristics becomes unavoidable. Our findings set high expectations for these heuristics, suggesting that even minor improvements in graph matching strategies can lead to significantly better generated graphs.

\bibliographystyle{unsrtnat}
\bibliography{references}

\end{document}